\documentclass[times,twocolumn,final,authoryear]{elsarticle}

\usepackage{ycviu}
\usepackage{framed,multirow}
\usepackage{amsmath}
\usepackage{amssymb}
\usepackage{latexsym}
\usepackage{hyperref}
\usepackage{bbm}
\usepackage{url}
\usepackage[table]{xcolor}
\usepackage{xcolor}
\definecolor{newcolor}{rgb}{.8,.349,.1}
\usepackage{subfigure}
\DeclareMathOperator*{\argmax}{argmax}
\journal{Computer Vision and Image Understanding}

\begin{document}

\thispagestyle{empty}

\clearpage
\thispagestyle{empty}
\ifpreprint
  \vspace*{-1pc}
\fi

\clearpage
\thispagestyle{empty}

\ifpreprint
  \vspace*{-1pc}
\else
\fi









\clearpage

\ifpreprint
  \setcounter{page}{1}
\else
  \setcounter{page}{1}
\fi

\begin{frontmatter}

\title{Domain Adaptation for Multi-label Image Classification: a Discriminator-free Approach}

\author[1]{Inder Pal \snm{Singh}\corref{cor1}} 
\cortext[cor1]{Corresponding author: 
  Tel.: +33 665958069;}
\ead{inder.singh@uni.lu}
\author[1,2]{Enjie \snm{Ghorbel}}
\author[1]{Anis \snm{Kacem}}
\author[1]{Djamila \snm{Aouada}}

\address[1]{Interdisciplinary Centre for Security, Reliability and Trust (SnT), University of Luxembourg, Luxembourg}
\address[2]{Cristal Laboratory, National School of Computer Sciences, University of Manouba, Tunisia}

\received{1 May 2013}
\finalform{10 May 2013}
\accepted{13 May 2013}
\availableonline{15 May 2013}
\communicated{S. Sarkar}

\begin{abstract}
This paper introduces a discriminator-free adversarial-based approach termed DDA-MLIC for Unsupervised Domain Adaptation (UDA) in the context of Multi-Label Image Classification (MLIC). While recent efforts have explored adversarial-based UDA methods for MLIC, they typically include an additional discriminator subnet. Nevertheless, decoupling the classification and the discrimination tasks may harm their task-specific discriminative power.
Herein, we address this challenge by presenting a novel adversarial critic directly derived from the task-specific classifier. Specifically, we employ a two-component Gaussian Mixture Model (GMM) to model both source and target predictions, distinguishing between two distinct clusters. Instead of using the traditional Expectation Maximization (EM) algorithm, our approach utilizes a Deep Neural Network  (DNN) to estimate the parameters of each GMM component. Subsequently, the source and target GMM parameters are leveraged to formulate an adversarial loss using the Fr\'echet distance. The proposed framework is therefore not only fully differentiable but is also cost-effective as it avoids the expensive iterative process usually induced by the standard EM method. 
The proposed method is evaluated on several multi-label image datasets covering three different types of domain shift. The obtained results demonstrate that DDA-MLIC outperforms existing state-of-the-art methods in terms of precision while requiring a lower number of parameters. The code is made publicly available at \url{github.com/cvi2snt/DDA-MLIC}.
\end{abstract}


\end{frontmatter}



\section{Introduction}
\label{sec:intro}

{M}{ulti}-label Image Classification  (MLIC) is an active research topic within the computer vision community, given its relevance in numerous applications such as object recognition~\cite{multi-object}, scene classification~\cite{deeply}, and attribute recognition~\cite{human-attr, deepfake}. Its primary objective is to predict the presence or absence of a predefined set of objects within a given image. 

\begin{figure}[!t]
\centering
\begin{tabular}{@{}c@{}}
    \includegraphics[width=0.99\linewidth]{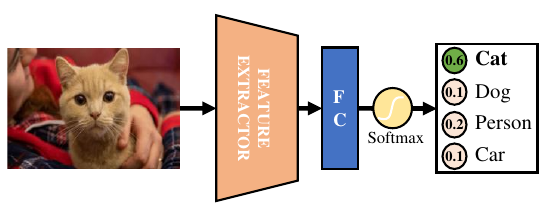} \\[\abovecaptionskip]
    \small (a) Single-label image classification
  \end{tabular}
  \begin{tabular}{@{}c@{}}
    \includegraphics[width=0.99\linewidth]{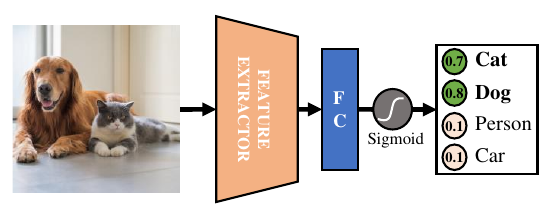} \\[\abovecaptionskip]
    \small (b) Multi-label image classification
  \end{tabular}
\hfill
\caption{The work of ~\cite{daln} cannot be directly applied to MLIC due to the differences between the two tasks~\cite{dda-mlic}: (a) Single-label image classification uses a softmax activation function to convert the predicted logits into probabilities such that the sum of all class probabilities is equal to one; and (b) on the other hand, multi-label image classification uses sigmoid activation where each logit is scaled between $0$ and $1$, giving higher probability values for the objects present in an image.}
\label{fig:difference}
\end{figure}



\noindent Thanks to the recent advancements in deep learning, several MLIC methods~\cite{ml-gcn, asl, ml-agcn, iml-gcn} have achieved remarkable performance on well-known benchmarks~\cite{coco,voc}. However, inheriting from the limitations of deep learning, existing MLIC methods are also negatively impacted by the \textit{domain shift} phenomenon. In other words, an MLIC method trained using data from a given domain, usually called \textit{source domain}, will suffer from degraded performance when tested on samples belonging to an unseen domain, referred to as \textit{target domain}. A direct solution is to simply label these target data and use them as additional training samples. Nevertheless, such a process is both resource-intensive and time-consuming. 

To handle this issue, \textit{Unsupervised Domain Adaptation (UDA)} methods have been proposed~\cite{uda, dann, mcd, mk-mmd} as an alternative. Instead of relying on annotated source data solely, UDA techniques take advantage of unlabelled target samples to minimize the shift between the source and the target domains. 

Existing UDA approaches have been primarily focusing on the problem of single-label image classification~\cite{dann, cada, mada, mk-mmd, mdd, mcd} and semantic segmentation~\cite{uda-sis1, uda-sis2, uda-sis3, uda-sis4}, giving less attention to other computer vision tasks including multi-label classification. Indeed, a limited number of UDA methods~\cite{da-maic,chest-xray, da-agcn} has been proposed for the specific case of multi-label image classification. These methods mainly take inspiration from adversarial UDA techniques for single-label image classification to implicitly reduce the domain shift. Similar to~\cite{dann}, these adversarial approaches integrate an additional domain discriminator coupled with a min-max two-player game. This strategy guides the generator to extract domain-invariant features that can fool the discriminator. However, as highlighted in~\cite{daln}, adopting such an adversarial training may cause mode collapse, resulting in a lower task-specific discriminative power.

To handle this issue in the context of single-label classification, Chen et al.~\cite{daln} proposed to reuse the classifier as a discriminator. More precisely, they introduced an adversarial critic based on the difference between inter-class and intra-class correlations of the classifier probability predictions.
However, unlike single-label classification, the per-class prediction probabilities in MLIC are not linearly dependent, thereby are not constrained to sum up to one, as depicted in Fig.~\ref{fig:difference}. Thus, a direct extension of the approach proposed in~\cite{daln} to MLIC is only possible by employing multiple binary classifiers, e.g., one for each class. In this way, the critic proposed in~\cite{daln} can be used by computing the correlations between the predicted probabilities of each binary classifier. Nevertheless, such an approach remains sub-optimal since the domain alignment is realized for each label classifier independently, disregarding the inter-class correlations. Our experiments in Section~\ref{sec:exp} support this hypothesis.

\begin{figure}[t]
\centering  
\begin{minipage}[b]{\linewidth}
  \label{fig:intuition_source}
  \centering
  \centerline{\includegraphics[width=6cm]{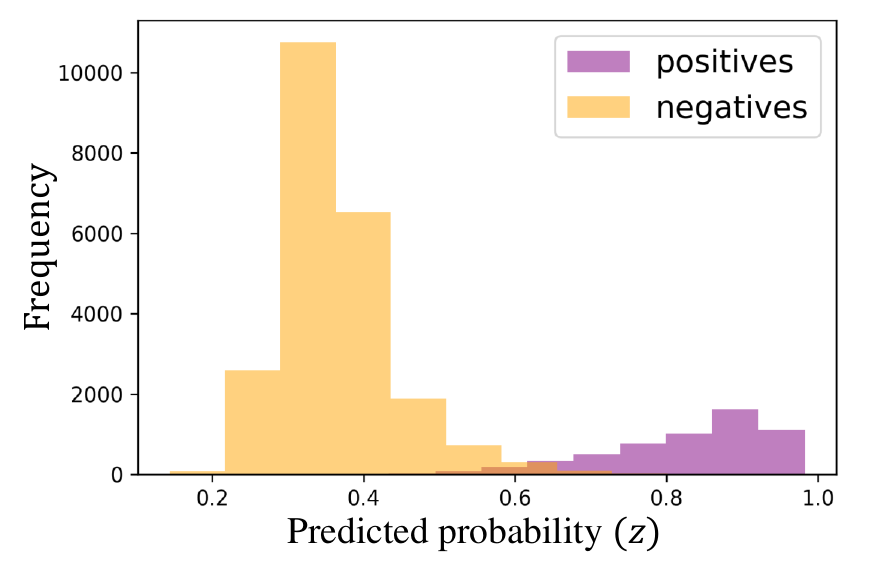}}
  \centerline{(a) Source $\rightarrow$ Source}\medskip
\end{minipage}
\begin{minipage}[b]{\linewidth}
\label{fig:intuition_target}
  \centering
  \centerline{\includegraphics[width=6cm]{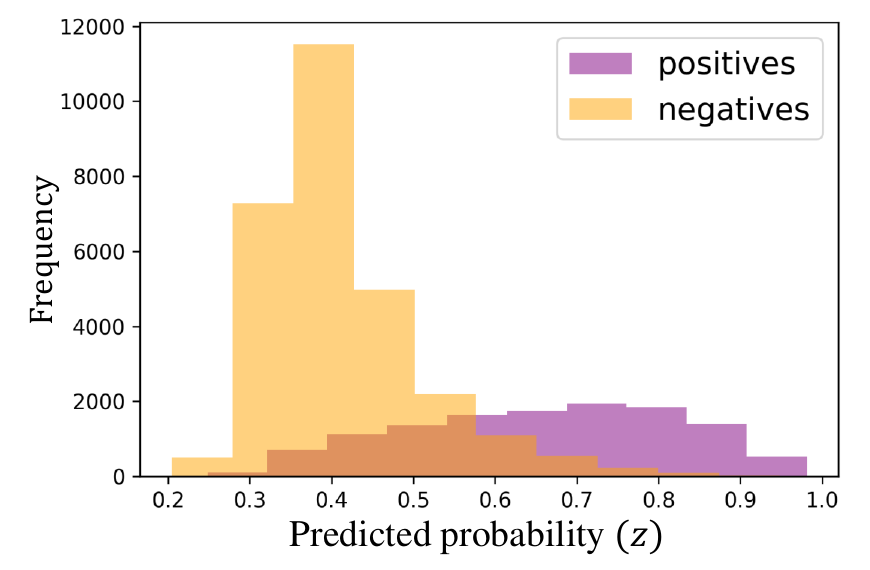}}
  \centerline{(b) Source $\rightarrow$ Target}\medskip
\end{minipage}
\hfill
\caption{Histogram of classifier predictions. Predicted probabilities using source-only trained classifier\protect\footnotemark[1] on: (a) source dataset\protect\footnotemark[2] $(\mathcal I_s)$, and (b) target dataset\protect\footnotemark[2] $(\mathcal I_t)$.}
\label{fig:intuition}
\end{figure}

In~\cite{dda-mlic}, we introduced a novel discriminator-free adversarial-based UDA method called DDA-MLIC, specifically tailored to MLIC. Motivated by~\cite{daln}, the task-specific classifier has been reused as a discriminator to avoid mode collapse. For that purpose, a novel critic suitable to the task of MLIC has been proposed. In particular, this critic is computed by clustering probability predictions into two sets (one in the neighborhood of 0 and another one in the neighborhood of 1), estimating their respective distributions and quantifying the distance between the estimated distributions from the source and target data.  The proposed idea is mainly inspired by the following observation: source samples tend to be classified (as positive or negative) more confidently than target ones, as illustrated in Fig.~\ref{fig:intuition}. The same figure also shows that the distribution of predictions is formed by two peaks; suggesting the suitability of a bimodal distribution model. Therefore, we argued that the distribution shape of probability predictions can implicitly enable the discrimination between source and target data. Practically, we proposed to fit a Gaussian Mixture Model (GMM) with two components on both the source and target predictions. Finally, a Fr\'echet distance~\cite{frechet} between the estimated pair of components has been employed for defining the introduced discrepancy measure. However, the use of the standard Expectation-Maximisation (EM) algorithm for estimating GMM introduces two main limitations, namely:

(1) \textbf{The non-differentiability}: the EM step is not differentiable as it breaks the chain rule. Hence, the EM does not contribute to the gradient-based optimization.

(2) \textbf{The demanding computational cost}: the EM algorithm is resource-intensive since it is based on iterative optimization. 

This paper presents a solution that extends the DDA-MLIC method to handle the two aforementioned limitations. Instead of relying on a non-differentiable and iterative traditional EM algorithm, the proposed method utilizes a neural block that mimics the EM optimization process. 
This block called \textit{DeepEM} is used for computing the GMM parameters based on a closed-form solution while ensuring the backpropagation of the related gradients through the entire network. As a result, only a single iteration is needed.
The experimental results show that the proposed approach outperforms state-of-the-art methods, including DDA-MLIC, in terms of mean Average Precision (mAP) while significantly reducing the average training time per batch and the number of network parameters.

This work is an extended version of~\cite{dda-mlic}. In summary, the additional contributions with respect to ~\cite{dda-mlic} are the following:

\begin{itemize}
\item A fully differentiable discriminator-free UDA for MLIC, based on the integrated DeepEM block.

    \item Additional theoretical details explaining the advantages of the proposed DeepEM over traditional EM algorithms for GMM fitting.
    \item A comprehensive and extended experimental analysis, demonstrating the superiority of the proposed method over state-of-the-art techniques in terms of mean Average Precision (mAP) and training time. 
\end{itemize}

The rest of the paper is organized as follows. Section~\ref{sec:related_works} discusses existing works on standard MLIC and UDA for multi-class and multi-label classification.
Section~\ref{sec:prob_motivation} formulates the problem of UDA in MLIC and details our motivation for reusing the classifier as a discriminator. Section~\ref{sec:method} introduces the discriminator-free approach DDA-MLIC and later details the proposed DeepEM block proposed as a replacement of the standard EM. The experimental analysis and discussion are detailed in Section~\ref{sec:exp}. 
Section~\ref{sec:lim} discusses the limitations of the proposed approach. 
Finally, Section~\ref{sec:conclusion} concludes this work and draws some interesting perspectives.

\footnotetext[1]{TResNet-M~\cite{tresnet} trained on UCM~\cite{ucm-ml}.}
\footnotetext[2]{Source: UCM~\cite{ucm-ml} validation set (420 images), Target: AID~\cite{aid-ml} validation set (600 images).}

\section{Related works}
\label{sec:related_works}
In this section, we start by reviewing the state-of-the-art on standard Multi-label Image Classification (MLIC). Then, we discuss related works in the general field of UDA. Lastly, we present the limited literature devoted to the topic of UDA for MLIC.

\subsection{Multi-label Image Classification (MLIC)}
In the literature, recent works in MLIC have benefited from the widespread availability of large-scale multi-label image datasets~\cite{coco, voc, clipart} and the proven success of deep Convolutional Neural Networks (CNN)~\cite{vgg, resnet}. For example,  Hypotheses-CNN Pooling (HCP)~\cite{hcp} has leveraged the predictions of multiple CNN architectures, such as AlexNet~\cite{alexnet} and VGG-16~\cite{vgg}, pre-trained on ImageNet~\cite{imagenet}. Recently, in~\cite{asl} authors have proposed Asymmetric Loss (ASL) that focuses more on the items present in the image (positive labels) than the ones that are absent (negative labels). ASL coupled with a recently introduced CNN architecture, named TResNet~\cite{tresnet}, has shown impressive performance for MLIC. Alternatively, ML-GCN~\cite{ml-gcn} employs a Graph Convolutional Network (GCN) to model the label correlations. Based on a similar strategy, ML-AGCN~\cite{ml-agcn} proposed to adaptively learn the label graph topology instead of heuristically defining it. 

Nevertheless, the performance of these methods is conditioned by the availability of large-scale annotated datasets. Notably, a significant drop in performance can be observed when applied to unseen domains~\cite{da-maic}.

\subsection{Unsupervised Domain Adaptation (UDA) for Single-label Image Classification}
UDA techniques have been proposed for enhancing the robustness of deep learning frameworks while avoiding costly labeling interventions. Most of UDA efforts have been dedicated to the task of multi-class classification. In particular, they have mainly followed two paradigms to reduce the disparity between source and target domains. The first paradigm explicitly aims to minimize the distance between the statistical moments of source and target distributions~\cite{etd, mdd, mcd, mk-mmd}. 
The second one makes use of an adversarial learning strategy~\cite{dann} in order to implicitly reduce the domain shift~\cite{mada, cada}. In general, these methods employ an additional domain discriminator that is meant to distinguish between source and target data. To generate domain-invariant features, an adversarial training strategy is adopted, where the goal is to fool the discriminator while maintaining an acceptable discriminative power.  

Despite their effectiveness, existing discriminator-based adversarial approaches may suffer from the problem of mode collapse, which typically occurs under adversarial training. In order to handle this challenge, a discriminator-free adversarial approach has been recently proposed in~\cite{dda-mlic}. Specifically, an adversarial critic based on the difference between inter-class and intra-class correlations of the probability predictions of the classifier is proposed. However, the predicted probabilities in Multi-Label Image Classification (MLIC) do not necessarily sum up to one, in contrast to single-label classification, as illustrated in Fig.~\ref{fig:difference}. Hence, as mentioned earlier, this approach can only be extended to MLIC by reformulating the problem as multiple binary classifications that might lead to sub-optimal results.

\subsection{Unsupervised Domain Adaptation (UDA) for Multi-label Image Classification (MLIC)}
As highlighted earlier,  only few UDA methods have been proposed for the specific case of MLIC~\cite{ml-anet,da-maic,da-agcn}. ML-ANet~\cite{ml-anet} follows a moment-matching strategy as they propose the use of Multi-Kernel Maximum Mean Discrepancies (MK-MMD) in a Reproducing Kernel Hilbert Space (RKHS). More recently, motivated by the progress made in adversarial-based UDA, attempts to generalize discriminator-based UDA methods to  MLIC have emerged. Specifically, DA-MAIC~\cite{da-maic} adopts a graph-based MLIC framework and couples it with a domain discriminator trained adversarially. Similarly, DA-AGCN~\cite{da-agcn} also follows a standard discriminator-based strategy, but injects an additional attention mechanism in the graph-based MLIC subnetwork. 

However, these discriminator-based methods are equally threatened by the mode collapse issue discussed in~\cite{daln}. Therefore, in this work, we propose a novel adversarial critic extracted from the task-specific classifier itself, thereby eliminating the use of an additional discriminator.

\section{Problem Formulation and Motivation}
\label{sec:prob_motivation}
In this section, we first formulate the problem of Unsupervised Domain Adaptation (UDA) for Multi-Label Image Classification (MLIC). Later, we detail our motivation behind reusing the multi-label classifier as a discriminator.

\subsection{Problem Formulation}
\label{sec:prob}
Let $\mathcal D_s= (\mathcal I_s, \mathcal Y_s)$ and $\mathcal D_t= (\mathcal I_t, \mathcal Y_t)$ be the source and target datasets, respectively, with $P_s$ and $P_t$ being their respective probability distributions such that $P_s \neq P_t$. Let us assume that they are both composed of $C$ object category labels. Note that $\mathcal I_s=\{\mathbf{I}_s^j\}_{j=1}^{n_s}$ and $\mathcal I_t=\{\mathbf{I}_t^j\}_{j=1}^{n_t}$  refer to the sets of $n_s$ source and $n_t$ target image samples, respectively, while $\mathcal Y_s=\{\mathbf{y}_s^j\}_{j=1}^{n_s}$ and $\mathcal Y_t=\{\mathbf{y}_t^j\}_{j=1}^{n_t}$ are their associated sets of labels. 

Let us denote by $\mathcal I$ the set of all images such that $\mathcal{I}~=~\mathcal{I}_s~\cup~\mathcal{I}_t$. Given an input image $\mathbf I \in \mathcal I $ with $\mathbf y\in \{0,1\}^C$ being its label, the goal of \textit{unsupervised domain adaptation} for \textit{multi-label image classification} is to estimate a function $f:\mathcal I \mapsto \{0,1\}^C$ such that,

\begin{equation}
    f(\mathbf I)=\mathbbm{1}_{f_c \circ f_g({\mathbf I})> \tau} = \mathbbm{1}_{\mathbf{Z}>\tau}=\mathbf y \ ,
\end{equation}

\noindent
where $f_g:~\mathcal{I}~\mapsto~\mathbb{R}^d$ extracts $d$-dimensional features, $f_c:~\mathbb R^d~\mapsto~[0,1]^C$ predicts the probability of object presence, $\mathbf{Z}=f_c\circ f_g (\mathbf I)\in [0,1]^C$ corresponds to the predicted probabilities, $\mathbbm{1}$ is an indicator function, $>$ is a comparative element-wise operator with respect to a chosen threshold  $\tau$. 
Note that only $\mathcal D_s$ and $\mathcal{I}_t$ are used for training.  In other words, the target dataset is assumed to be unlabeled.

To achieve this goal, some existing methods~\cite{da-maic} have adopted an adversarial strategy by considering an additional discriminator $f_d$ that differentiates between source and target data. Hence, the model is optimized using a classifier loss $\mathcal L_{cls}$ such as the asymmetric loss (ASL)~\cite{asl} and an adversarial loss $\mathcal L_{adv}$ defined as,
\begin{equation}
\begin{split}
\label{eq:domain_loss}
    \mathcal L_{adv} = \mathbb{E}_{f_{g}(\mathbf I_s) \sim  \bar{P}_s}
 \log \frac{1}{f_d(f_g(\mathbf I_s))} + \\ \mathbb{E}_{f_{g}(\mathbf I_t) \sim \bar{P}_t}  \log \frac{1}{(1-f_d(f_g(\mathbf I_t))} \ ,
 \end{split}
\end{equation}
where $\bar{P}_s$ and $\bar{P}_t$ are the distributions of the learned features from source and target samples $\mathcal I_s$ and $\mathcal{I}_t$, respectively.

While the adversarial paradigm has shown great potential~\cite{da-maic}, the use of an additional discriminator $f_d$ which is decoupled from $f_c$ may lead to mode collapse as discussed in~\cite{daln}.  Inspired by the same work, we aim at addressing the following question -- \textit{Could we leverage the outputs of the task-specific classifier $f_c \circ f_g$ in the context of multi-label classification for implicitly discriminating the source and the target domains?}

\subsection{Motivation: Domain Discrimination using the Distribution of the Classifier Output}
\label{sec:critic}
The goal of MLIC is to identify the classes that are present in an image (\emph{i.e.}, \textit{positive labels}) and reject the ones that are absent (\emph{i.e.}, \textit{negative labels}). Hence, the classifier $f_c$ is expected to output high probability values for the positive labels and low probability values for the negative ones. Formally, let $z=\theta (f_c(f_g(\mathbf I))) =\theta(\mathbf{Z} )\sim \hat P$ be the random variable modelling the predicted probability of any class and $\hat P$ its probability distribution, with $\theta$ being a uniform sampling function that returns the predicted probability of a randomly selected class. In general, a well-performing classifier is expected to classify confidently both negative and positive samples. Ideally, this would mean that the probability distribution $\hat P$ should be formed by two clusters with low variance in the neighborhood of $0$ and $1$, respectively denoted by $\mathcal C_0$ and $\mathcal C_1$. Hence, our hypothesis is that a drop in the classifier performance due to a domain shift can be reflected in $\hat P$. 

Let $z_s=\theta (f_c(f_g(\mathbf I_s)))$ $\sim \hat{P_s}$ and $z_t=~\theta (f_c(f_g(\mathbf I_t)))~\sim~\hat P_t$ be the random variables modelling the predicted probability obtained from the source and target data and $\hat P_s$ and $\hat P_t$ be their distributions, respectively. Concretely, we propose to investigate whether the shift between the source and target domains is translated in  $\hat P_s$ and $\hat P_t$. If a clear difference is observed between $\hat P_s$ and $\hat P_t$,  this would mean that the classifier $f_c$ should be able to discriminate between source and target samples. Thus, this would allow the definition of a suitable critic directly from the classifier predictions.

To support our claim, we trained a model\protect\footnotemark[4] $f$
 using the labelled source data $\mathcal{D}_s$ without involving the target images\protect\footnotemark[5] $\mathcal{I}_t$. In Fig.\ref{fig:intuition}~(a), we visualize the histogram of the classifier probability outputs when the model is tested on the source domain. It can be clearly observed that the predicted probabilities on the source domain, denoted by $\mathbf{z}_s$, can be grouped into two separate clusters.  Fig.\ref{fig:intuition}~(b) shows the same histogram when the model is tested on target samples. In contrast to the source domain, the classifier probability outputs, denoted by $\mathbf{z}_t$, are more spread out in the target domain. In particular, the two clusters are less separable than in the source domain. This is due to the fact that the classifier $f_c$ benefited from the supervised training on the source domain, and as a result it gained an implicit discriminative ability between the source and target domains. 

Motivated by the observations discussed above, we propose to reuse the classifier to define a critic function based on $\hat P_s$ and $\hat P_t$. In what follows, we describe our approach including the probability distribution modelling ($\hat P_s$ and $\hat P_t$) and the adversarial strategy for domain adaptation.   

\footnotetext[3]{TResNet-M~\cite{tresnet} trained on UCM~\cite{ucm-ml}.}
\footnotetext[4]{Source: UCM~\cite{ucm-ml} validation set (420 images), Target: AID~\cite{aid-ml} validation set (600 images).}

\begin{figure}[!t]
\centering
\begin{minipage}[b]{0.48\linewidth}
  \centering
\centerline{\includegraphics[width=4.1cm]{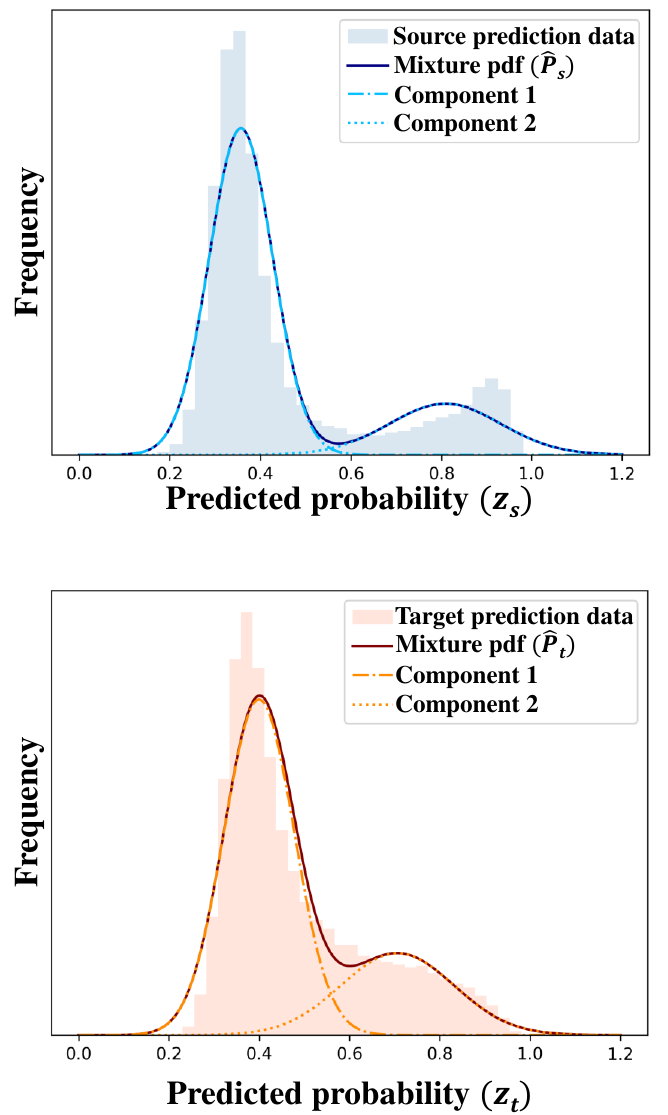}}
  \centerline{(a) GMM fitting.}
\end{minipage}
\begin{minipage}[b]{0.48\linewidth}
  \centering
  \centerline{\includegraphics[width=4.1cm]{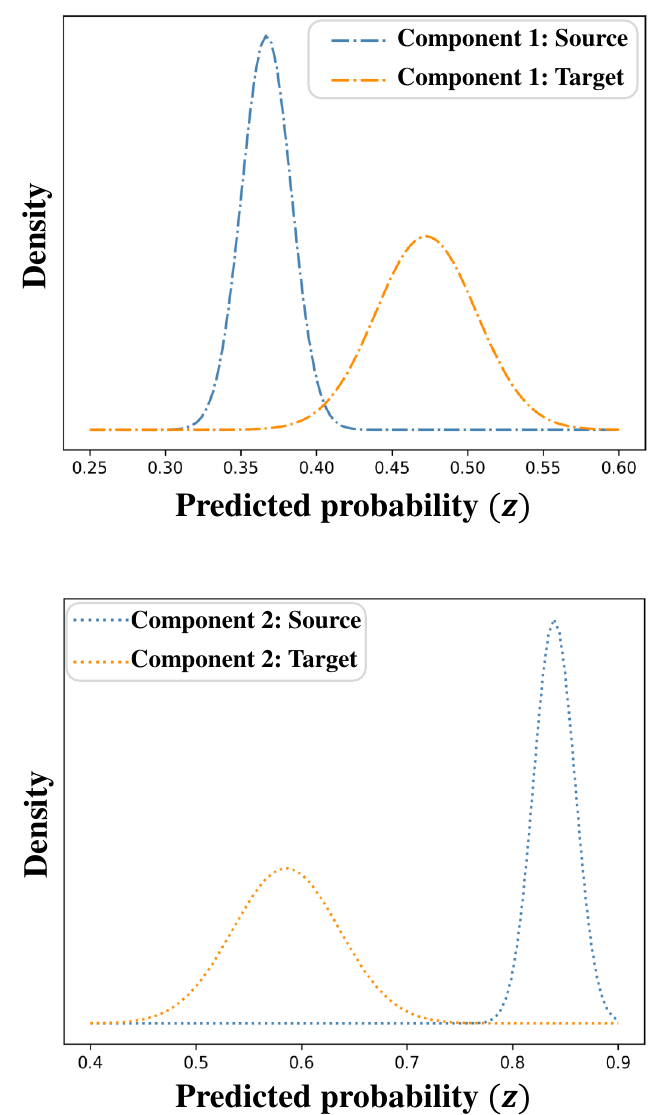}}
  \centerline{(b) Gaussians of components.}
\end{minipage}
\hfill
\caption{(a) The classifier\protect\footnotemark[4] predictions $\mathbf{z}_s$ and $\mathbf{z}_t$ for both source and target datasets\protect\footnotemark[5], respectively, can be grouped into two clusters. Hence, a two-component GMM can be fitted for both source ($\hat{P}_s$) and target ($\hat{P}_t$). While the first component is close to 0, the second is close to 1, (b) A component-wise comparison between source ($\hat P_s^1, \hat P_s^2$) and target ($\hat P_t^1, \hat P_t^2$) Gaussians of distributions extracted from the fitted GMM confirms that target predictions are likely to be farther from 0 and 1 with a higher standard deviation than the source.}

\label{fig:intro} 
\end{figure}
\begin{figure*}[!t]
  \centering
  \centerline{\includegraphics[width=0.95\linewidth]{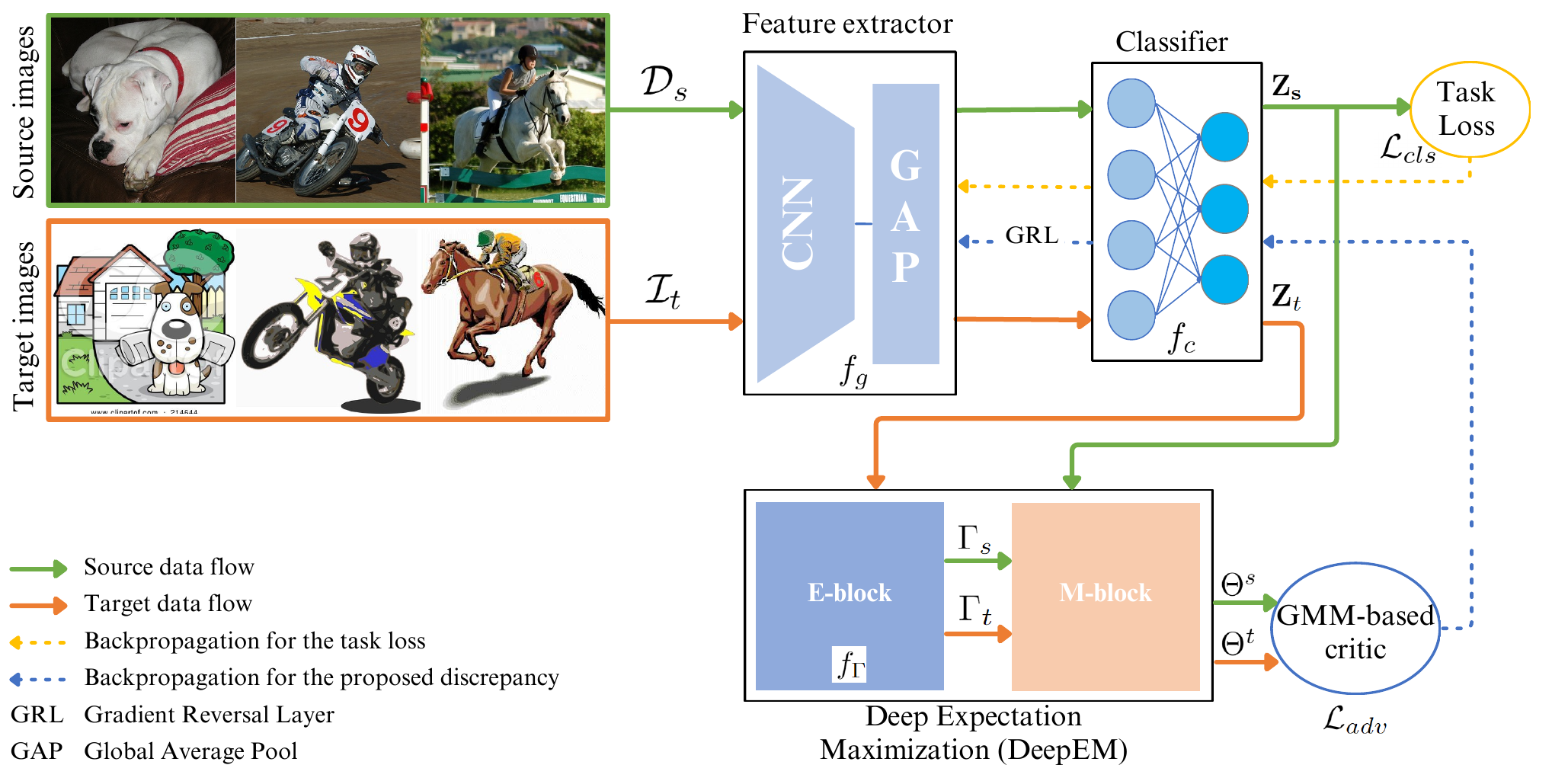}}
\caption{The overall architecture of DDA-MLIC with the proposed DeepEM block consists of the following components: The feature extractor ($f_g$) learns discriminative features from both source and target images. The task classifier ($f_c$) performs two actions simultaneously: 1) it learns to accurately classify source samples using a supervised task loss $\mathcal{L}_{cls}(\mathcal{D}_s)$, and 2) when acting as a discriminator, it aims to minimize the proposed GMM-based discrepancy $\mathcal{L}_{adv}(\mathcal{D}_s, \mathcal{I}_t)$ between source $(\mathbf{z}_s)$ and target $(\mathbf{z}_t)$ predictions using the proposed DeepEM block, while $f_g$ simultaneously works to maximize it.}
\label{fig:arch}
\end{figure*}

\section{Proposed Approach}

\label{sec:method}

In this section, we first detail the DDA-MLIC approach, including the novel adversarial critic derived from the task classifier using the standard EM algorithm. Later, we introduce the proposed DeepEM block that overcomes the limitations of the traditional EM, namely, non-differentiability with respect to the overall architecture and a high computational cost. Finally, we provide an overview of the proposed architecture incorporating an end-to-end learning process for multi-label prediction and unsupervised discriminator-free domain adaptation.

\subsection{DDA-MLIC: an Implicit Adversarial Critic from the Task Classifier}
\label{sec:adv_critic}
As discussed in Section~\ref{sec:critic}, the classifier probability predictions are usually formed by two clusters with nearly Gaussian distributions. Consequently, as shown in Fig.~\ref{fig:intro} (a), we suggest approximating the two distributions $\hat P_s$ and $\hat P_t$ by a two-component Gaussian Mixture Model (GMM) defined in Eq.~\eqref{eq:2A} as follows,

\begin{equation}
\label{eq:gmm_s}
   \hat P_s(\mathbf{z}_s) \approx \sum_{k=1}^2 \pi_k^s \mathcal{N}(\mathbf{z}_s|\mu_k^s, \sigma_k^s) \ ,
\end{equation} 
and, 
\begin{equation}
\label{eq:gmm_t}
   \hat P_t(\mathbf{z}_t) \approx \sum_{k=1}^2 \pi_k^t \mathcal{N}(\mathbf{z}_t|\mu_k^t, \sigma_k^t) \ ,
\end{equation} 

\noindent where $\mathcal{N}(\mathbf{z}_t|\mu_k^t, \sigma_k^t)$ denotes the $k$-th Gaussian distribution, with the mean  $\mu_k^t$ and the standard deviation $\sigma_k^t$, fitted on the target predicted probabilities $\mathbf{z}_t$ and $\pi_k^t$ its mixture weight such that $\pi_1^t+\pi_2^t=1$. Similarly,  $\mathcal{N}(\mathbf{z}_s|\mu_k^s, \sigma_k^s)$ denotes the $k$-th Gaussian distribution, with the mean $\mu_k^s$ and the standard deviation $\sigma_k^s$, fitted on the source predicted probabilities $\mathbf{z}_s$ and $\pi_k^s$ its mixture weight such that $\pi_1^s+\pi_2^s=1$. For estimating the two GMM models, the standard EM algorithm is used.

In both the source and target domains, we posit that the initial component of the Gaussian Mixture Model (GMM) aligns with the cluster $\mathcal C_0$ (featuring a mean in proximity to 0), while the second component corresponds to $\mathcal C_1$ (with a mean in proximity to $1$). However, due to a large number of negative predictions as compared to positive ones, the component $\mathcal C_0$ tends to be more dominant. In fact, in a given image, only few objects are usually present from the total number of classes. To alleviate this phenomenon, in DDA-MLIC, we proposed to extract two Gaussian components from the source and target GMM, ignoring the estimated weights illustrated in Fig.\ref{fig:intro} (b).

In order to reuse the classifier as a discriminator in the context of UDA for MLIC, in DDA-MLIC~\cite{dda-mlic}, we proposed to redefine the adversarial loss $(\mathcal{L}_{adv})$ by computing a Fr\'echet distance~$(d_\mathrm{F})$~\cite{frechet} between each pair of the estimated source and target components, using Eq.~\eqref{eq:m_intro} and~\eqref{eq:m_step}, from a given cluster as follows,

\begin{equation}
\label{eq:loss-adv}
    \mathcal{L}_{adv} = \sum_{k=1}^2  \alpha_k d_\mathrm{F}(\mathcal{N}(\mathbf{z}_t|\mu_k^t, \sigma_k^t),\mathcal{N}(\mathbf{z}_s|\mu_k^s, \sigma_k^s)) \ ,
\end{equation}

\noindent with $\alpha_k$ weights that are empirically fixed. 
Since the computed distributions are univariate Gaussians, the Fréchet distance between two distributions, also called the 2-Wasserstein (2W) distance, is chosen as it can be explicitly  computed as follows,

\begin{equation}
    d^2_\mathrm{F}(\mathcal N(\mathbf{z}_1|\mu_1, \sigma_1),\mathcal N(\mathbf{z}_2|\mu_2, \sigma_2)) = 
    (\mu_1-\mu_2)^2+(\sigma_1-\sigma_2)^2,
\end{equation}
where $\mathcal N(\mathbf{z}_1|\mu_1, \sigma_1)$ and $ \mathcal N(\mathbf{z}_2|\mu_2, \sigma_2)$ are two Gaussians with a mean of $\mu_1$ and $\mu_2$ and a standard deviation of $\sigma_1$ and $\sigma_2$, respectively. In addition, compared to the commonly used 1-Wasserstein (1W) distance, it considers second-order moments. Finally, in~\cite{arjovsky2017wasserstein}, the 2W distance has been demonstrated to have nicer properties e.g., continuity and differentiability,  for optimizing neural networks as compared to other divergences and distances between two distributions such as the Kullback-Leibler  (KL) divergence and the Jensen-Shannon (JS) divergence. The relevance of the 2W distance is further discussed in Section~\ref{sec:distance}. 

\subsection{Deep Expectation Maximization (DeepEM)}
\label{sec:diff_em}
Due to its iterative nature, the standard Expectation-Maximization (EM) algorithm is computationally demanding. Furthermore, the GMM fitting step based on EM is non-differentiable with respect to the DDA-MLIC architecture. As a result, it does not impact the backward propagation, posing a challenge to the overall learning process. 
Alternatively, we propose to use an additional block termed Deep Expectation Maximization (DeepEM), inspired by~\cite{deep-gmr}. The proposed DeepEM consists of two blocks: 1) a Multi-layer Perceptron (MLP) network called \textit{E-block}, denoted as $f_{\Gamma}$, and 2) a parameter-free computational block, called \textit{M-block}.

\subsubsection{E-block}

Let $\mathbf{Z} = \{\mathbf{z}_1, \mathbf{z}_2,...,\mathbf{z}_B\}$ represents the predicted probabilities for any given batch formed by $B$ images, where $\mathbf{z}_i \in \mathbb R^C$ is the predicted probabilities vector for sample $i$. Given these predicted probabilities as input, the E-block outputs the responsibility matrix denoted as $\Gamma \in \mathbb R^{BC\times2}$ such that,

\begin{equation}
\begin{split}
    \Gamma &= f_\Gamma(\mathbf{Z})\\ &= [\hat{\gamma}_{ik}]_{i\in\{1,BC\},k\in\{1,2\}},
\end{split}
\end{equation}
where $\hat{\gamma}_{ik}$ is the responsibility estimated by the network.

In summary, the E-block replaces the E-step of the standard EM algorithm. By employing this DNN-based strategy, we overcome two challenges: 1) the chain rule is not broken allowing the differentiability of the E-step,  2) the iterative and resource-intensive computation of the responsibilities is not required. Instead, this can be achieved in a single iteration.

\subsubsection{M-block}
The M-block relies on the latent variable formulation of the traditional EM algorithm, utilizing the closed-form maximum likelihood updates for GMM parameters conditioned on the soft responsibilities $(\Gamma)$ produced by the E-block. More details on this formulation are provided in the Appendix. Let us denote the source and target responsibilities by $\Gamma_s$ and $\Gamma_t$, respectively resulting from the E-block $(f_\Gamma)$. Given the predicted probabilities $\mathbf{z}_s$ and $\mathbf{z}_t$ for the source and target samples, respectively, the M-block computes the GMM parameters $\Theta^s_k=(\pi^s_k, \mu^s_k, \sigma^s_k)$ and $\Theta_k^t=(\pi^t_k, \mu^t_k, \sigma^t_k)$ for $k\in\{1,2\}$ using the closed-form solution depicted in Eq.~\eqref{eq:m_step} in the Appendix. The proposed adversarial critic defined in Eq.~\eqref{eq:loss-adv} can be directly applied as follows,

\begin{equation}
\label{eq:loss_adv_new}
    \mathcal{L}_{adv} = \sum_{k=1}^2\alpha_k \left( (\mu^s_k-\mu^t_k)^2 + (\sigma^s_k-\sigma^t_k)^2 \right),
\end{equation}
\noindent
where $\alpha_k$ is a hyperparameter that regulates the contributions from each GMM component. The variation of this hyperparameter $\alpha$ is discussed in Section~\ref{sec:sensistivity}.

\subsection{Overall Architecture}
The overall architecture of the proposed approach is illustrated in Fig.~\ref{fig:arch}. Similar to~\cite{daln}, the proposed DDA-MLIC consists of: (1) a feature extractor $(f_g)$ that aims to extract discriminative image features from both source $(\mathcal{I}_s)$ and target $(\mathcal{I}_t)$ images, and (2) a classifier $(f_c)$ that performs the multi-label classification and at the same time implicitly discriminates between source and target data. 

When acting as a classifier, $f_c$ aims to minimize the supervised classification loss~\cite{asl} $(\mathcal{L}_{cls})$ using the annotated source dataset $\mathcal{D}_s$. However, when operating as a discriminator, the output of $f_c$ is fed to the proposed DeepEM block. It returns the estimated source and target GMMs, i.e., $\Theta^s$ and $\Theta^t$, respectively, thereby enabling the computation of the proposed adversarial loss $\mathcal{L}_{adv}$ as reformulated in Eq.~\eqref{eq:loss_adv_new}.
A Gradient Reversal Layer (GRL) between $f_g$ and $f_c$ enforces the feature extractor to fool the classifier when acting as a discriminator, thereby implicitly learning domain-invariant features. Consequently, the network is trained by engaging in a min-max game as depicted below,

\begin{equation}
    \min_{f_c}\max_{f_{g}} \mathcal{L}_{adv}.
\end{equation}

In summary, the overall loss function used to train DDA-MLIC is described below,

\begin{equation}
    \min_{f_g, f_c} \left\{\mathcal{L}_{cls}(\mathcal{D}_s) + \lambda \max_{f_g} \mathcal{L}_{adv}(\mathcal{D}_s, \mathcal{I}_t)\right\},
\end{equation}

\noindent
where $\lambda$ is another hyper-parameter that weights $\mathcal{L}_{cls}$ and $\mathcal{L}_{adv}$. 

\section{Experiments}
\label{sec:exp}
In this section, we report the performed experiments and discuss the obtained results. First, we present the datasets used for our experimental study. Later, we detail the experimental settings as well as the implementation details. Finally, we report and analyze the obtained results.

\subsection{Experimental Settings}
\subsubsection{Datasets}
In our experiments, different types of domain gaps are considered, namely, (1) the domain shift due to the use of different sensors, (2) the domain gap existing between simulated and real data (3) the discrepancy resulting from different weather conditions. Due to the limited availability of cross-domain multi-label datasets, we convert several object detection and semantic segmentation datasets to the task of MLIC. 

\paragraph{Cross-sensor domain shift}
Similar to~\cite{da-maic}, we use three multi-label aerial image datasets that have been captured using different sensors resulting in different resolutions, pixel densities and altitudes, namely: 1) the \textbf{AID}~\cite{aid-ml} multi-label dataset was created from the original multi-class AID dataset~\cite{aid} by labeling $3000$ aerial images, including $2400$ for training and $600$ for testing, with a total of $17$ categories. 2) the \textbf{UCM}~\cite{ucm-ml} multi-label dataset was recreated from the original multi-class classification dataset~\cite{ucm} with a total of $2100$ image samples containing the same $17$ object labels as AID. We randomly split the dataset into training and testing sets with 2674 and 668 image samples, respectively. 3) the \textbf{DFC}~\cite{dfc} multi-label dataset provides $3342$ high-resolution images with training and testing splits of, respectively, $2674$ and $668$ samples labeled from a total of $8$ categories. In our experiments, the $6$ common categories between DFC and the other two benchmarks are used.

\paragraph{Sim2real domain shift}
We use the following two datasets to investigate the domain gap between real and synthetic scene understanding images. 1) \textbf{PASCAL-VOC}\cite{voc} is one of the most widely used real image datasets for MLIC with more than $10K$ image samples. It covers $20$ object categories. The training and testing sets contain $5011$ and $4952$ image samples, respectively. 2) \textbf{Clipart1k}~\cite{clipart} provides 1000 synthetic clipart image samples, annotated with 20 object labels, similar to VOC. Since it is proposed for the task of object detection, we make use of the category labels for bounding boxes to create a multi-label version. Half of the data are used for training and the rest is used for testing. 

\paragraph{Cross-weather domain shift}
In order to study the domain shift caused by different weather conditions, two widely used urban street datasets have been used, namely:
1) \textbf{Cityscapes}~\cite{cityscapes} which is introduced for the task of semantic image segmentation and consists of $5000$ real images captured in the daytime. 2) \textbf{Foggy-cityscapes}~\cite{foggy} which is a synthesized version of Cityscapes where an artificial fog is introduced. We generate a multi-label version of these datasets for the task of MLIC considering only $11$ categories out of the original $19$ to avoid including the objects that appear in all the images. 

\subsubsection{Implementation Details}
The proposed work makes use of TResNet-M~\cite{tresnet} as a backbone and the Asymmetric Loss (ASL)~\cite{asl} as the task loss. All the methods are trained using the Adam optimizer with a cosine decayed maximum learning rate of $10^{-3}$. For all the experiments, we make use of NVIDIA TITAN V with a batch size of $64$ for a total of $25$ epochs or until convergence. The input image resolution has been fixed to  $224\times224$.

\subsubsection{Baselines}
To evaluate our approach, we categorize methods into three groups. \textit{MLIC} methods are trained solely on source datasets and evaluated directly on target datasets without any domain adaptation strategy, including both direct and indirect approaches. \textit{Disc.-based} and \textit{Disc.-free} methods utilize both labeled source and unlabeled target datasets, with and without the domain discriminator, respectively and are evaluated on the target dataset. We adapt the discriminator-free approach DALN~\cite{daln}, originally designed for UDA in single-label classification, to MLIC by computing the adversarial critic for multiple binary predictions.

\subsubsection{Evaluation Metrics}
Similar to~\cite{dda-mlic}, we report several metrics including the number of model parameters (\# params), mean Average Precision (mAP), average per-Class Precision (CP), average per-Class Recall (CR), average per-Class F1-score (CF1), average Overall Precision (OP), average overall recall (OR) and average Overall F1-score (OF1). We consider seven datasets: AID, UCM, DFC, VOC, Clipart, Cityscapes, and Foggycityscapes, resulting in seven experimental settings: AID $\rightarrow$ UCM, UCM $\rightarrow$ AID, AID $\rightarrow$ DFC, UCM $\rightarrow$ DFC, VOC $\rightarrow$ Clipart, Clipart $\rightarrow$ VOC, and Cityscapes $\rightarrow$ Foggy. For example, AID $\rightarrow$ UCM indicates that AID is fixed as the source dataset during training while UCM is the target dataset. The reported results are based on the testing set of the target dataset.

\subsection{Quantitative Analysis}

\begin{table*}[t]
\caption{Cross-sensor domain shift: comparison with the state-of-the-art in terms of number of model parameters (in millions), and \% scores of mAP, per-class averages (CP, CR, CF1) and overall averages (OP, OR, OF1) for aerial image datasets. Two settings are considered, \emph{i.e.}, AID $\rightarrow$ UCM and UCM $\rightarrow$ AID. Best results are highlighted in \textbf{bold}.}
\label{table:aid_ucm_results_both}
\resizebox{\textwidth}{!}{%
\begin{tabular}{|l|l|c|ccccccc|ccccccc|}
\hline
           \multirow{2}{*}{\textbf{Type}} & \multirow{2}{*}{\textbf{Method}}       & \multirow{2}{*}{\textbf{\# params}}                 & \multicolumn{7}{c|}{\textbf{AID $\rightarrow$ UCM}}                & \multicolumn{7}{c|}{\textbf{UCM $\rightarrow$ AID}}   \\
           \cline{4-17}
           
& & & \textbf{mAP}                   & \textbf{P\_C}                  & \textbf{R\_C}                  & \textbf{F\_C}                  & \textbf{P\_O}                  & \textbf{R\_O}                  & \textbf{F\_O}                  & \textbf{mAP}                   & \textbf{P\_C}                  & \textbf{R\_C}                  & \textbf{F\_C}                  & \textbf{P\_O}                  & \textbf{R\_O}                  & \textbf{F\_O}                  \\
\hline
           & ResNet101~\cite{resnet}          & 42.5            & 57.5& \textbf{60.0} & 47.5& 47.0& 69.1& 71.5& \textbf{70.3} & 51.7& 50.6& 29.6& 33.9& 88.0& 48.5& 62.5\\
           & ML-GCN~\cite{ml-gcn}             & 44.9            & 53.7& 55.3& 44.3& 45.9& 70.2& 68.7& 69.4& 51.3& 50.1& 29.9& 34.0& 88.0& 49.7& 63.6\\
           
           & ML-AGCN~\cite{ml-agcn}            & 36.6            & 55.2& 36.6& \textbf{64.9} & 45.1& 45.0& \textbf{88.1} & 59.6& 52.1& 48.2& \textbf{47.4} & \textbf{42.9} & 77.1& \textbf{79.8} & \textbf{78.4} \\
\multirow{-4}{*}{MLIC}          & ASL (TResNetM)~\cite{asl}     & 29.4            & 55.4& 48.7& 52.8& 47.1& 58.7& 79.1& 67.4& 54.1& 54.5& 40.2& 41.9& 85.4& 65.1& 73.9\\
     \hline
           & DANN (TResNetM + ASL)~\cite{dann}                   & 29.4            & 52.5& 59.1& 31.6& 36.3& \textbf{70.9} & 53.7& 61.1& 51.6& 52.1& 23.2& 27.9& 83.2& 27.8& 41.7\\
      
\multirow{-2}{*}{Disc-based}    & DA-MAIC (TResNetM+ASL)~\cite{da-maic}                  & 36.6            & 54.4& 55.3& 37.5& 38.6& 68.0& 67.9& 67.9& 50.5& 51.8& 22.9& 29.0& 91.6& 35.2& 50.8\\
\hline
           & DALN (TResNetm + ASL)~\cite{daln}                   & 29.4            & 53.1& 53.3& 32.4& 36.7& 69.2& 53.9& 60.6& 53.2& 52.2& 29.3& 32.7& 82.0& 41.2& 54.8\\
           \cline{2-17}
           & \textbf{DDA-MLIC (OURS)}    & 29.4            & \textbf{63.2} & 52.5& 63.7& \textbf{55.1} & 59.4& 82.8& 69.2& 54.9& 53.9& 30.4& 35.5& 84.6& 41.0& 55.3\\

\multirow{-3}{*}{Disc-free}     & \textbf{DDA-MLIC with DeepEM (OURS)} & 29.4            & 60.5& 53.5& 51.8& 48.8& 61.4& 76.8& 68.2& \textbf{56.2} & \textbf{58.0} & 19.0& 26.5& \textbf{97.4} & 31.5& 47.6                 \\
\hline
\end{tabular}}
\end{table*}

\begin{table*}[t]
\caption{Cross-sensor domain shift: comparison with the state-of-the-art in terms of number of model parameters (in millions), and \% scores of mAP, per-class averages (CP, CR, CF1) and overall averages (OP, OR, OF1) for aerial image datasets. Two settings are considered, \emph{i.e.}, AID $\rightarrow$ DFC and UCM $\rightarrow$ DFC. Best results are highlighted in \textbf{bold}.}
\label{table:dfc_all}
\resizebox{\textwidth}{!}{%
\begin{tabular}{|l|l|c|ccccccc|ccccccc|}
\hline
           &              &                 & \multicolumn{7}{c}{\textbf{AID $\rightarrow$ DFC}}                & \multicolumn{7}{|c|}{\textbf{UCM $\rightarrow$ DFC}}                \\
           \cline{4-17}
\multirow{-2}{*}{\textbf{Type}} & \multirow{-2}{*}{\textbf{Method}} & \multirow{-2}{*}{\textbf{\# params}} & \textbf{mAP}                   & \textbf{P\_C}                  & \textbf{R\_C}                  & \textbf{F\_C}                  & \textbf{P\_O}                  & \textbf{R\_O}                  & \textbf{F\_O}                  & \textbf{mAP}                   & \textbf{P\_C}                  & \textbf{R\_C}                  & \textbf{F\_C}                  & \textbf{P\_O}                  & \textbf{R\_O}                  & \textbf{F\_O}                  \\
\hline
           & ResNet101~\cite{resnet}    & 42.5            & 56.9& 52.9& 61.5& 48.7& 46.1& 63.7& 53.5& 66.4& 74.4& 31.2& 36.9& 67.2& 37.2& 47.9\\
           & ML-GCN~\cite{ml-gcn}       & 44.9            & 58.9& \textbf{56.7} & 57.9& 45.8& 45.7& 65.0& 53.7& 64.6& 72.4& 32.0& 35.6& 64.4& 38.9& 48.5\\
           & ML-AGCN~\cite{ml-agcn}      & 36.6            & 51.6& 41.5& \textbf{83.8} & 52.3& 40.2& \textbf{88.7} & 55.3& 70.3& 68.4& \textbf{56.1} & 47.8& 53.8& \textbf{58.5} & 56.0\\
\multirow{-4}{*}{MLIC}          & ASL (TResNetM)~\cite{asl}                    & 29.4            & 56.1& 49.6& 68.4& 49.9& 43.5& 74.1& 54.8& 68.9& 66.3& 53.1& 44.0& 52.6& 57.0& 54.7\\
\hline
           & DANN (TResNetM + ASL)~\cite{dann}             & 29.4            & 43.0& 40.7& 13.6& 19.3& 46.0& 15.6& 23.3& 64.1& 77.3& 22.6& 30.1& 68.6& 26.5& 38.2\\
\multirow{-2}{*}{Disc-based}    & DA-MAIC (TResNetM+ASL)~\cite{da-maic}            & 36.6            & 55.4& 49.8& 60.4& 44.7& 47.3& 64.1& 54.4& 65.8& 71.4& 39.3& 39.7& 59.9& 44.6& 51.1\\
\hline
           & DALN (TResNetm + ASL)~\cite{daln}             & 29.4            & 44.7& 43.7& 23.8& 27.6& 48.9& 27.4& 35.1& 65.6& \textbf{82.6} & 21.3& 32.0& \textbf{75.2} & 22.1& 34.1\\
           \cline{2-17}
           & \textbf{DDA-MLIC (OURS) }                  & 29.4            & 62.1& 47.6& 75.5& \textbf{55.3} & 48.9& 76.2& \textbf{59.6} & 70.6& 67.2& 55.7& \textbf{49.3} & 55.0& 58.4& \textbf{56.6} \\
           
\multirow{-3}{*}{Disc-free}     & \textbf{DDA-MLIC with DeepEM (OURS) }                  & 29.4            & \textbf{63.2} & 50.7& 50.9& 42.7& \textbf{50.7} & 56.3& 53.4& \textbf{73.1} & 74.9& 49.5& 47.7& 63.1& 51.0& 56.4\\
\hline
\end{tabular}}
\end{table*}

\begin{table*}[t]
\caption{Sim2Real domain shift: comparison with the state-of-the-art in terms of number of model parameters (in millions), and \% scores for mAP, per-class averages (CP, CR, CF1) and overall averages (OP, OR, OF1) for scene understanding datasets. Two settings are considered, \emph{i.e.}, VOC $\rightarrow$ Clipart and Clipart $\rightarrow$ VOC. Best results are highlighted in \textbf{bold}.}
\label{table:voc_clipart}
\resizebox{\textwidth}{!}{%
\begin{tabular}{|l|l|c|ccccccc|ccccccc|}
\hline
           &                    &                 & \multicolumn{7}{c}{\textbf{VOC $\rightarrow$ Clipart}}            & \multicolumn{7}{|c|}{\textbf{Clipart $\rightarrow$ VOC}}            \\
         \cline{4-17}
\multirow{-2}{*}{\textbf{Type}} & \multirow{-2}{*}{\textbf{Method}}       & \multirow{-2}{*}{\textbf{\# params}} & \textbf{mAP}                   & \textbf{P\_C}                  & \textbf{R\_C}                  & \textbf{F\_C}                  & \textbf{P\_O}                  & \textbf{R\_O}                  & \textbf{F\_O}                  & \textbf{mAP}                   & \textbf{P\_C}                  & \textbf{R\_C}                  & \textbf{F\_C}                  & \textbf{P\_O}                  & \textbf{R\_O}                  & \textbf{F\_O}                  \\
\hline

           & ResNet101~\cite{resnet}          & 42.5            & 38.0& 64.8& 14.3& 22.5& 82.3& 18.3& 29.9& 50.1& 66.2& 17.5& 25.5& 83.9& 29.6& 43.7\\
           & ML-GCN~\cite{ml-gcn}             & 44.9            & 43.5& 62.5& 20.3& 28.4& 86.6& 27.8& 42.1& 43.1& 57.9& 21.0& 26.8& 73.5& 30.6& 43.2\\
           & ML-AGCN~\cite{ml-agcn}            & 36.6            & 53.7& 75.5& 35.5& 44.4& 79.1& 39.9& 53.1& 38.0& 45.5& 25.1& 28.2& 61.8& 36.6& 45.9\\
\multirow{-4}{*}{MLIC}          & ASL (TResNetM)~\cite{asl}     & 29.4            & 56.8& 72.0& 38.5& 47.6& 82.8& 45.7& 58.9& 64.2& 69.0& 30.7& 37.3& 80.0& 45.7& 58.2\\
\hline
           & DANN (TResNetM + ASL)~\cite{dann}                   & 29.4            & 47.0& 77.0& 22.0& 32.5& 86.8& 23.6& 37.1& 67.0& 76.8& 23.3& 32.6& \textbf{93.1} & 20.4& 33.4\\
\multirow{-2}{*}{Disc-based}    & DA-MAIC (TResNetM+ASL)~\cite{da-maic}                  & 36.6            & \textbf{62.3} & 77.4& \textbf{42.6} & \textbf{51.6} & 83.1& \textbf{51.0} & \textbf{63.2} & 74.3& 84.5& 53.9& 63.0& 83.7& 57.7& 68.3\\
\hline
           & DALN (TResNetm + ASL)~\cite{daln}                   & 29.4            & 45.0& 82.2& 21.4& 32.6& 92.0& 22.7& 36.4& 66.7& 78.3& 22.2& 31.7& 90.8& 18.0& 30.0\\
           \cline{2-17}
           & \textbf{DDA-MLIC (OURS) }   & 29.4            & 61.4& \textbf{84.7} & 28.1& 39.4& 90.9& 33.3& 48.8& 77.0& 86.9& 29.3& 38.2& 88.4& 35.3& 50.4\\
           
\multirow{-3}{*}{Disc-free}     & \textbf{DDA-MLIC with DeepEM (OURS) } & 29.4            & 62.0& 80.8& 23.4& 34.6& \textbf{94.8} & 25.4& 40.0& \textbf{82.8} & \textbf{88.6} & \textbf{57.0} & \textbf{65.8} & 86.4& \textbf{58.8} & \textbf{70.0}\\
\hline
\end{tabular}}
\end{table*}

\begin{table*}[t]
\caption{Cross-weather domain shift: comparison with the state-of-the-art in terms of number of model parameters (in millions), and \% scores of mAP, per-class averages (CP, CR, CF1) and overall averages (OP, OR, OF1) for urban street datasets. Cityscapes $\rightarrow$ Foggy is the setting that is considered. Best results are highlighted in \textbf{bold}.}
\label{table:coco_city}
\centering
\resizebox{0.65\textwidth}{!}{%
\begin{tabular}{|l|l|c|ccccccc|}
\hline
           &              &                 & \multicolumn{7}{c|}{\textbf{Cityscapes $\rightarrow$ Foggy}}       \\
           \cline{4-10}
\multirow{-2}{*}{\textbf{Type}} & \multirow{-2}{*}{\textbf{Method}} & \multirow{-2}{*}{\textbf{\# params}} & \textbf{mAP}                   & \textbf{P\_C}                  & \textbf{R\_C}                  & \textbf{F\_C}                  & \textbf{P\_O}                  & \textbf{R\_O}                  & \textbf{F\_O}                  \\
\hline
           & ResNet101~\cite{resnet}    & 42.5            & 58.2& 53.6& 27.8& 32.2& \textbf{93.2} & 48.3& 63.7\\
           & ML-GCN~\cite{ml-gcn}       & 44.9            & 56.6& 56.1& 34.6& 38.8& 89.4& 56.9& 69.6\\
           & ML-AGCN~\cite{ml-agcn}      & 36.6            & 60.7& 60.1& 48.3& 50.9& 81.7& \textbf{71.2} & \textbf{76.1} \\
\multirow{-4}{*}{MLIC}          & ASL (TResNetM)~\cite{asl}                    & 29.4            & 61.3& 66.7& \textbf{50.8} & \textbf{53.8} & 79.2& 70.5& 74.6\\
\hline
           & DANN (TResNetM + ASL)~\cite{dann}             & 29.4            & 53.5& 50.6& 12.5& 18.6& 89.5& 21.8& 35.1\\
\multirow{-2}{*}{Disc-based}    & DA-MAIC (TResNetM+ASL)~\cite{da-maic}            & 36.6            & 61.9& 70.7& 37.2& 42.7& 90.2& 59.6& 71.7\\
\hline
           & DALN (TResNetm + ASL)~\cite{daln}             & 29.4            & 54.8& 56.8& 9.5 & 25.4& 90.2& 33.8& 49.2\\
           \cline{2-10}
           & \textbf{DDA-MLIC (OURS) }                  & 29.4            & 62.3& \textbf{73.7} & 45.7& 48.9& 84.1& 69.3& 76.0\\
           
\multirow{-3}{*}{Disc-free}     & \textbf{DDA-MLIC with DeepEM (OURS)}                    & 29.4            & \textbf{63.2} & 71.9& 43.2& 45.4& 85.8& 67.3& 75.5    \\
\hline
\end{tabular}}
\end{table*}

\subsubsection{Comparison with state-of-the-art methods}
Table~\ref{table:aid_ucm_results_both}, Table~\ref{table:dfc_all}, Table~\ref{table:voc_clipart} and Table~\ref{table:coco_city} quantitatively compare the proposed approach to state-of-the-art methods.  It can be seen that our model requires an equal or fewer number of parameters than other state-of-the-art works, with a total number of $29.4$ million parameters. We achieve the best performance in terms of mAP for AID $\rightarrow$ UCM, UCM $\rightarrow$ AID, AID $\rightarrow$ DFC, UCM $\rightarrow$ DFC, Clipart $\rightarrow$ VOC and Cityscapes $\rightarrow$ Foggy. 

The first four rows of Table~\ref{table:aid_ucm_results_both}, Table~\ref{table:dfc_all}, Table~\ref{table:voc_clipart} and Table~\ref{table:coco_city} report the obtained results using different methods of MLIC without DA~\cite{resnet,ml-gcn,ml-agcn,asl}. It can be observed that our method consistently outperforms all these methods under all settings (cross-sensor, sim2Real, and cross-weather) in terms of mAP showing the effectiveness of the proposed DA method for MLIC. 

Furthermore, the results reported in the $5^{th}$ and $6^{th}$ rows of Table~\ref{table:aid_ucm_results_both}, Table~\ref{table:dfc_all}, Table~\ref{table:voc_clipart}, Table~\ref{table:coco_city} show that the proposed discriminator-free DA method clearly outperforms discriminator-based DA approaches for MLIC~\cite{dann,da-maic} on cross-sensor and cross-weather domain shift settings in terms of mAP. This observation does not hold for the sim2Ream domain shift, where our approach records an mAP improvement of $8.5\%$ over other discriminator-based approaches on Clipart~$\rightarrow$~VOC setting, but wslightly surpasses with $0.3\%$ in terms of mAP DA-MAIC~\cite{da-maic} on VOC $\rightarrow$ Clipart setting.

We also compare our method to the discriminator-free method proposed in~\cite{daln} for single-label DA and adapted to MLIC as stated in Section~\ref{sec:related_works}. Unsurprisingly, our method outperforms the adapted version of DALN for MLIC under all settings, reaching an improvement of more than $16\%$ in terms of mAP on the Clipart~$\rightarrow$~VOC and VOC~$\rightarrow$~Clipart scheme.

\subsubsection{DeepEM versus traditional EM}
The last two rows of Table~\ref{table:aid_ucm_results_both}, Table~\ref{table:dfc_all}, Table~\ref{table:voc_clipart} and Table~\ref{table:coco_city} report the results using the proposed DDA-MLIC without and with DeepEM. The adoption of a differentiable EM strategy showcases a substantial performance improvement under the three settings. It is worth highlighting that the mAP score is improved by approximately 6\% in Sim2Real domain shift for Clipart $\rightarrow$ VOC with the proposed DeepEM block. This further supports the relevance of the proposed differentiable approach.

\begin{table*}[!t]
\centering
\caption{Ablation study: the mAP metric is reported. w/o w/: refers to without and with respectively.  Best results are highlighted in \textbf{bold}.}
\label{table:ablation}
\resizebox{0.8\textwidth}{!}{%
\begin{tabular}{|l|c|c|c|c|c|c|}
\hline
\textbf{Methods} & \textbf{AID$\rightarrow$UCM} & \textbf{UCM$\rightarrow$AID} & \textbf{AID$\rightarrow$DFC} & \textbf{UCM$\rightarrow$DFC} & \textbf{VOC$\rightarrow$Clipart} & \textbf{Clipart$\rightarrow$VOC} \\
\hline
\textbf{Ours w DeepEM}    &60.52	&56.23	&63.23	&73.06	&61.97	&82.80

        \\
 \textbf{Ours w/o DeepEM} & 63.24 (+2.7)      & 54.90 (-1.4)     & 62.13 (-1.1)     & 70.64 (-2.4)     & 61.44 (-0.5)         & 76.96 (-5.8)         \\
\hline
Ours w/o DA      & 55.45 \textbf{(-5.1)}      & 54.12 \textbf{(-2.1)}     & 56.09 \textbf{(-7.1)}     & 68.91 \textbf{(-4.1)}     & 56.78 \textbf{(-5.2)}         & 64.15 \textbf{(-18.7)}         \\
\hline
Ours w/ Discr.   & 52.54 \textbf{(-8.0)}     & 51.60 \textbf{(-4.6)}     & 51.60 \textbf{(-11.6)}     & 64.06 \textbf{(-9.0) }    & 46.97 \textbf{(-15.0)}         & 67.03 \textbf{(-15.8)}          \\
\hline
\end{tabular}}
\end{table*}

\begin{table*}[!t]
\caption{mAP comparison of the proposed EM-based GMM clustering with k-means clustering.}
\label{table:kmeans}
\centering
\resizebox{0.8\textwidth}{!}{%
\begin{tabular}{|l|c|c|c|c|c|c|}
\hline
\textbf{Methods} & \textbf{AID$\rightarrow$UCM} & \textbf{UCM$\rightarrow$AID} & \textbf{AID$\rightarrow$DFC} & \textbf{UCM$\rightarrow$DFC} & \textbf{VOC$\rightarrow$Clipart} & \textbf{Clipart$\rightarrow$VOC} \\
\hline
\textbf{Ours w DeepEM} &60.52	&56.23	&63.23	&73.06	&61.97	&82.80
         \\
\hline
Ours (with k-means)        & 53.58 \textbf{(-6.9)}      & 52.20 \textbf{(-4.0)}     & 58.46 \textbf{(-4.8)}      & 68.06 \textbf{(-5.0) }    & 49.24 \textbf{(-12.7) }              & 68.27 \textbf{(-14.5)}              \\
\hline
\end{tabular}}
\end{table*}

\begin{table*}[!t]
\caption{Performance of the proposed method in terms of mAP using the  KL-divergence and the 1-Wasserstein (1W) distance.  Best results are highlighted in \textbf{bold}.}
\label{table:kl}
\centering
\resizebox{0.8\textwidth}{!}{%
\begin{tabular}{|l|c|c|c|c|c|c|}
\hline
\textbf{Methods}        & \textbf{AID$\rightarrow$UCM} & \textbf{UCM$\rightarrow$AID} & \textbf{AID$\rightarrow$DFC} & \textbf{UCM$\rightarrow$DFC} & \textbf{VOC$\rightarrow$Clipart} & \textbf{Clipart$\rightarrow$VOC} \\
\hline
\textbf{Ours w DeepEM} &60.52	&56.23	&63.23	&73.06	&61.97	&82.80
         \\
\hline
Ours (with KL)               & 56.44 \textbf{(-4.1)}      & 53.51 \textbf{(-2.7)}     & 53.17 \textbf{(-10.1)}     & 64.55 \textbf{(-8.5)}     & 52.62 \textbf{(-9.3) }        & 77.86 \textbf{(-4.9)}     \\
Ours (with 1W) & 53.60 \textbf{(-6.9)} & 53.20 \textbf{(-3.0)} & 57.80 \textbf{(-5.4)} & 69.70 \textbf{(-3.4)} & 60.50 \textbf{(-1.5)} & 75.50 \textbf{(-7.3)}\\
\hline
\end{tabular}}
\end{table*}

\subsubsection{Training time}
\begin{figure}[!t]
  \centering
  \centerline{\includegraphics[width=\linewidth]{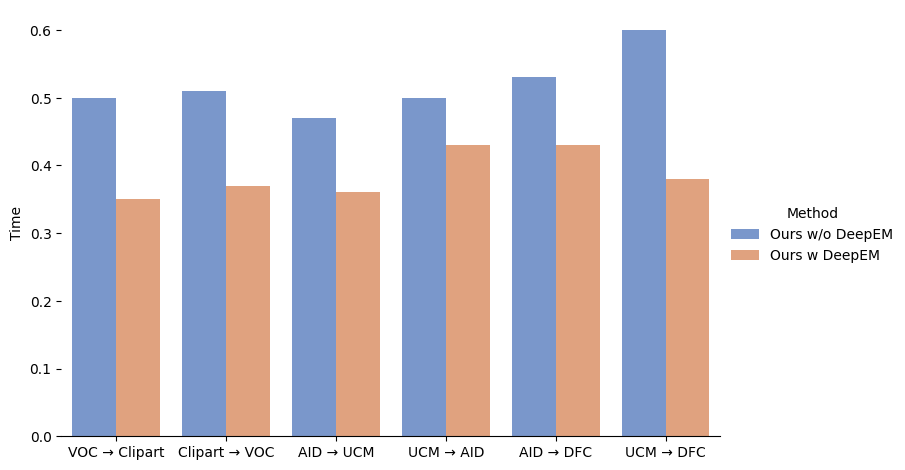}}
\caption{Comparison of the average training time per batch with and without DeepEM.}
\label{fig:batch_time}
\end{figure}

In order to showcase the efficiency of the proposed DeepEM, Fig.~\ref{fig:batch_time} compares the average training time needed to process one batch of source and target images using DDA-MLIC, with and without DeepEM. The figure shows that by replacing the traditional iterative EM process with an appropriate deep neural network significantly reduces the training time.

\begin{table*}[!t]
\caption{Sensitivity analysis:  performance of the proposed approach in terms of mAP (\%) when varying $\lambda_1$ and $\lambda_2$.  Best results are highlighted in \textbf{bold}.}
\label{table:sensitivity}
\centering
\resizebox{0.9\textwidth}{!}{%
\begin{tabular}{|c|c|c|c|c|c|c|c|}
\hline
\multirow{2}{*}{\textbf{$\alpha$ values ($\alpha_1, \alpha_2$)}} & \multicolumn{4}{c|}{\textbf{Cross-sensor}} & \multicolumn{2}{c|}{\textbf{Sim2real}} & \textbf{Cross-weather} \\
\cline{2-8}
 & \textbf{AID$\rightarrow$UCM} & \textbf{UCM$\rightarrow$AID} & \textbf{AID$\rightarrow$DFC} & \textbf{UCM$\rightarrow$DFC} & \textbf{VOC$\rightarrow$Clipart} & \textbf{Clipart$\rightarrow$VOC} & \textbf{City$\rightarrow$Foggy}  \\
\hline
$\alpha_1$=0.1, $\alpha_2$=0.9 & 55.86 & 54.05 & 60.29 & 71.43 & 82.55& 56.80& 61.66        \\
$\alpha_1$=0.2, $\alpha_2$=0.8 & 55.67 & \textbf{56.23} & 60.16 & 70.99 & 80.46& 56.61& 61.34      \\
$\alpha_1$=0.3, $\alpha_2$=0.7 & 56.00 & 54.20 & \textbf{63.23} & \textbf{73.06} & 81.92& 58.48& \textbf{63.23}      \\
$\alpha_1$=0.4, $\alpha_2$=0.6 & 55.57 & 55.89 & 60.65 & 72.84 & 81.29& 57.71& 61.57       \\
$\alpha_1$=0.5, $\alpha_2$=0.5 & 57.92 & 55.00 & 60.91 & 71.33 & \textbf{82.80} & \textbf{61.67} & 61.80                  \\
$\alpha_1$=0.6, $\alpha_2$=0.4 & 54.85 & 55.72 & 61.22 & 70.19 & 81.86& 58.28& 62.95      \\
$\alpha_1$=0.7, $\alpha_2$=0.3 & 58.35 & 54.44 & 61.51 & 71.85 & 82.26& 57.52& 60.59        \\
$\alpha_1$=0.8, $\alpha_2$=0.2 & \textbf{60.52} & 55.46 & 58.96 & 70.78 & 81.39& 58.03& 61.86      \\
$\alpha_1$=0.9, $\alpha_2$=0.1 & 58.21 & 54.14 & 58.42 & 71.40 & 81.54& 57.83& 60.91       \\
$\alpha_1$=1.0, $\alpha_2$=0.0 & 58.46 & 54.83 & 58.47 & 72.15 & 81.87& 59.94& 61.85       \\
\hline
\end{tabular}}
\end{table*}

\begin{figure}[!t]
  \centering
  \centerline{\includegraphics[width=\linewidth]{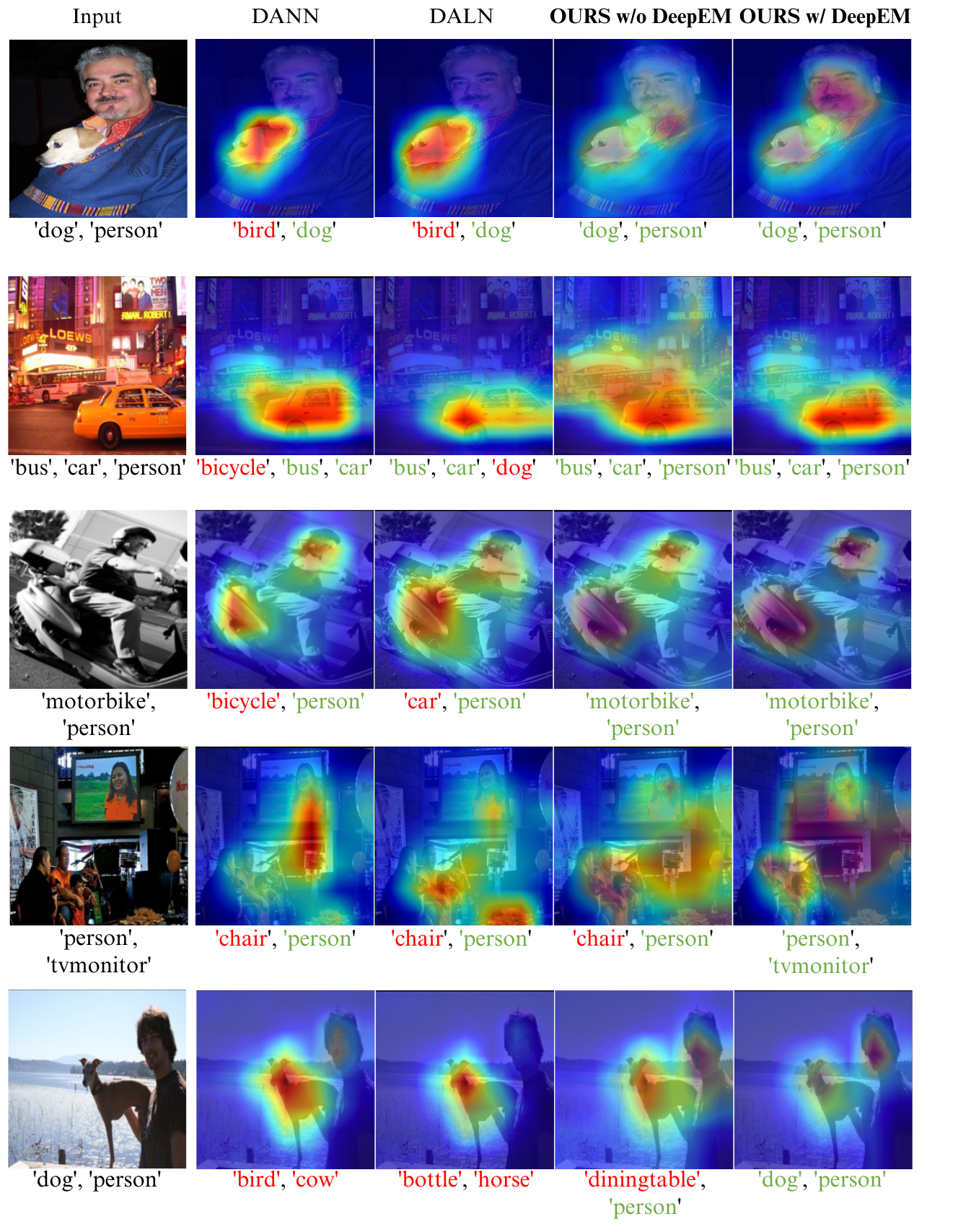}}
\caption{Qualitative analysis:  Heatmap visualization of the proposed approach, existing discriminator-based (DANN~\cite{dann}), and discriminator-free (DALN~\cite{daln}) methods. The first column exhibits some input images with their ground truth labels, while the next columns display the heatmaps generated by the considered methods, with predicted labels highlighted in green (if correct) and red (if incorrect).}
\label{fig:hm}
\end{figure}

\subsubsection{Ablation Study}
The results of the ablation study are presented in Table~\ref{table:ablation}.  We report the obtained mAP for the following settings, \emph{i.e.}, AID $\rightarrow$ UCM, UCM $\rightarrow$ AID, UCM $\rightarrow$ AID, UCM $\rightarrow$ DFC, VOC $\rightarrow$ Clipart and Clipart $\rightarrow$ VOC. 
The initial two rows display the mAP results obtained using the proposed approach with \textit{(w)} and without \textit{(w/o)} the inclusion of the DeepEM. The third row illustrates the mAP score in the absence of any domain adaptation strategy. The final row shows the results achieved when employing an adversarial domain adaptation approach, utilizing a conventional domain discriminator. Clearly, leveraging the classifier as a discriminator leads to a noticeable improvement in classification performance when dealing with a domain shift.

\subsection{Qualitative Analysis}
Fig.~\ref{fig:hm} presents a qualitative comparison between the proposed method and existing approaches. Specifically, we compare the proposed discriminator-free approach (with and without DeepEM) to both existing discriminator-based (DANN~\cite{dann}) and discriminator-free (DALN~\cite{daln}) methods. For a set of image samples, we visualize the Gradient-weighted Class Activation Mapping (Grad-CAM)~\cite{grad-cam} for all the aforementioned methods, along with the corresponding predicted labels. It can be noted that compared to our method both DANN and DALN fail to precisely activate the regions incorporating the present objects, hence leading to an incorrect prediction of the labels. Furthermore, the relevance of the differentiable EM-based strategy is illustrated in the last two columns of Fig.~\ref{fig:hm}. In these examples, the DeepEM-based network allows focusing more precisely on the corresponding objects, as compared to the standard EM, thereby enabling the correct prediction of the different categories present in the displayed images.

\subsection{Additional Analysis}

\subsubsection{Sensitivity analysis}
\label{sec:sensistivity}
In Table~\ref{table:sensitivity}, we report the variation in performance when varying the values $\alpha_1 $ and $\alpha_2 $, defined in Eq.~\eqref{eq:loss-adv}. More specifically, we report the mAP score achieved by DDA-MLIC with DeepEM for ten different combinations of $\alpha_1 $ and $\alpha_2 $ for the three types of domain shift. We start by giving smaller weights to the first GMM component (negative labels) and higher importance to the second one(positive labels). In most cases, especially under the cross-sensor and cross-weather domain shifts, we observe that assigning more weights to the positive component yields a better mAP score. This is in line with the reasoning behind the widely used Asymmetric Loss (ASL)~\cite{asl}, which focuses more on the positive labels than negative ones. For the Sim2real domain shift, however, assigning equal weights to both components provides the best mAP score.

\subsubsection{GMM versus k-means}
To justify the choice of GMM as a clustering technique, we compare it with another popular non-probabilistic clustering technique, known as k-means. In contrast to k-means, which uses hard thresholding to assign data points to specific clusters, GMM employs soft thresholding by maximizing the likelihood. Table~\ref{table:kmeans} compares the mAP scores obtained using the two methods. It is evident that using k-means results in a significant performance drop for all benchmarks.

\subsubsection{Distance and divergence measure analysis}
\label{sec:distance}
As mentioned in Section~\ref{sec:adv_critic}, the proposed adversarial critic, directly derived from the task-specific classifier, is based on the 2-Wasserstein distance (denoted as 2W) between the estimated source and target GMM components. 
To demonstrate the effectiveness of the 2W distance over other popular distances, we report ,in Table~\ref{table:kl}the mAP scores when integrating the KL-divergence and the 1-Wasserstein distance in our framework. The results clearly demonstrate that utilizing the 2W distance as a discrepancy measure outperforms other distances, thanks to its continuity and differentiability properties. 
Specifically, using the KL divergence or the 1-W distance as a discrepancy measure results in a slight to significant reduction in mAP across all benchmarks, ranging from 1.5\% to 10\%.

\section{Limitations}
\label{sec:lim}
While the proposed method has demonstrated superior performance across nearly all benchmarks, it is essential to acknowledge certain limitations. Specifically, in the case of the AID$\rightarrow$UCM benchmark (see Table~\ref{table:aid_ucm_results_both}), we observe that, under the cross-sensor domain shift, the proposed DDA-MLIC with DeepEM did not yield a noticeable performance improvement when compared to its counterpart based on the traditional EM. 
Furthermore, in the case of the sim2real domain shift, particularly in the VOC$\rightarrow$Clipart benchmark (see Table~\ref{table:voc_clipart}), the performance of the discriminator-free method did not surpass the discriminator-based approach DA-MAIC~\cite{da-maic}. We assume that this drop in performance might be due to the absence of an additional graph-based subnet, which is a crucial component for explicitly modeling label correlations. This limitation highlights the need for investigating additional strategies for modeling the label correlations. Moreover, our approach assumes that the categories that are present in source and target images are identical. Hence, it would be interesting to investigate in future works more challenging scenarios such as open-set unsupervised domain adaptation, where the source and target datasets might include non-common labels.

\section{Conclusion}
\label{sec:conclusion}
In this paper, a discriminator-free UDA approach for MLIC has been proposed. Unlike existing methods that employ an additional discriminator trained adversarially, our method utilizes the task-specific classifier to implicitly discriminate between source and target domains. This strategy aims to enforce learning domain-invariant features, while avoiding mode collapse. To achieve this, we redefine the adversarial loss using a Fr\'echet distance between the corresponding Gaussian Mixture Model (GMM) components estimated from the classifier probability predictions. 
A DNN-based Deep Expectation Maximization (DeepEM) is proposed to estimate the parameters of the GMM for ensuring differentiability and avoiding a costly iterative optimization.  Experiments conducted on several benchmarks encoding different domain shifts demonstrated that the proposed approach achieves state-of-the-art performance, while reducing the need for cumbersome architectures.
 
\newpage

\section*{Acknowledgments}
This research was funded in whole, or in part, by the Luxembourg National Research Fund (FNR), grant references BRIDGES2020/IS/14755859/MEET-A/Aouada and BRIDGES2021/IS/16353350/FaKeDeTeR. For the purpose of open access, and in fulfillment of the obligations arising from the grant agreement, the author has applied a Creative Commons Attribution 4.0 International (CC BY 4.0) license to any Author Accepted Manuscript version arising from this submission.





\bibliographystyle{model2-names}
\bibliography{refs}


\newpage

\section*{Appendix}
\label{sec:appendix}
The GMM is a mixture model that is formed by $K$ Gaussian components, as depicted in the following equation,

\begin{equation}
    P(\mathbf{x}|\Theta) = \sum_{k=1}^K \pi_k \mathcal{N}(\mathbf{x}|\mu_k,\Sigma_k),
    \label{eq:2A}
\end{equation}
\noindent
where $\mathbf{x}=\{x_1, x_2,...,x_N\}$ is an $N$-dimensional continuous-valued data vector (i.e. observations or features). The tuple $\Theta_k = \{\pi_k, \mu_k, \Sigma_k\},$ is formed by $\pi_k$, $\mu_k$ and $\Sigma_k$ which denote the mixture weights, the mean vector and the covariance matrix of the $k^{\text{th}}$ GMM component, respectively, with $\sum_{k=1}^K \pi_k = 1$. 

The Maximum Likelihood Estimation (MLE) is a common approach for estimating the mixture parameters $\Theta =~\{\Theta_k\}_{k \in \{\ 1,2,...,K\} } $ by maximizing the log-likelihood of the observations, given by,
\begin{equation}
\label{eq:llhood}
    l(\Theta | \mathbf{x}) =  \log\:P(\mathbf{x}|\Theta).
\end{equation}


Since directly maximizing $l(\Theta|\mathbf{x})$ is intractable, the EM algorithm maximizes instead a lower bound of $  l(\Theta | \mathbf{x})$ defined as, 

 \begin{equation}
     Q = \log \left\{ \sum_{\mathcal{C}} P(\mathcal{C}|\mathbf{x}, \Theta) \right\} .
 \end{equation}

 \noindent where $\mathcal{C}$ refers to a set of discrete latent variables.


 An iterative optimization of this bound alternates between two steps: in the E-step, a new estimate of the posterior probability distribution over $\mathcal{C}$ is computed, given the estimation of the parameters $\Theta$ denoted as $\Theta^{m}$ from the previous iteration $m$.
 In the context of GMM, the set $\mathcal{C}$ is defined as a set of binary latent variables $\mathcal{C}=(c_{ik})_{i \in \{1, ...,N\}, \newline k \in \{1, ...,K\}}$ for simplifying the optimization process. This helps in calculating the responsibility of each component $k$ in the mixture as follows,
 \begin{equation}
           \gamma_{ik}(\Theta^{m}) = Q(c_{ik}=1|\Theta_k^{m},x_i)
\label{eq:resp}
 \end{equation}

In the M-step, the parameters $\Theta$ are updated in order to maximize the expected log-likelihood using the posteriors computed in the E-step such that,

\begin{equation}
\label{eq:m_intro}
    \Theta^{m+1} =\argmax_{\Theta^{m}} Q(\Theta^{m+1}|\Theta^{m}).
    \end{equation}
\noindent

Thanks to the introduction of the binary latent variables $\mathcal{C}$, the computation of the updated parameters in Eq.~\eqref{eq:m_intro} can be done using a closed-form solution as detailed below,

\begin{equation}
\begin{split}
\label{eq:m_step}
 \pi_k^{m+1} &= \frac{1}{N} \sum_{i=1}^N {\gamma_{ik}}, \;\;\;\;
    \mu_k^{m+1} = \frac{\sum_{i=1}^N {\gamma}_{ik}x_i}{\sum_{i=1}^N {\gamma}_{ik}},\\
    \Sigma_k^{m+1} &= \frac{\sum_{i=1}^N {\gamma}_{ik}(x_i-\mu_k^{m})(x_i-\mu_k^{m})^T}{\sum_{i=1}^N {\gamma}_{ik}}.
\end{split}    
\end{equation}

The algorithm refines the values of the estimated parameters iteratively until convergence. Hence, this optimization process remains computationally costly.  
\end{document}